\documentclass[sigconf]{acmart}
\usepackage{makecell}
\usepackage{multirow}
\usepackage{bbding}
\usepackage{CJKutf8}
\usepackage{xcolor,soul}

\setcopyright{none}
\renewcommand\footnotetextcopyrightpermission[1]{}
\usepackage{caption}
\AtBeginDocument{%
  }

\begin{document}

\title{Tyee: A Unified, Modular, and Fully-Integrated Configurable Toolkit for Intelligent Physiological Health Care}

\author{Tao Zhou}
\authornote{These authors contributed equally to this research.}
\affiliation{%
  \institution{Hunan University}
  \city{Changsha}  
  \country{China}
}
\email{zhooutao@hnu.edu.cn}

\author{Lingyu Shu}
\authornotemark[1]
\affiliation{
  \institution{Hunan University}
  \city{Changsha}
  \country{China}
}
\email{shulingyu@hnu.edu.cn}

\author{Zixing Zhang}
\authornote{Zixing Zhang and Jing Han are corresponding authors.}
\affiliation{
  \institution{Hunan University}
  \city{Changsha}
    \country{China}
}
\email{zixingzhang@hnu.edu.cn}

\author{Jing Han}
\authornotemark[2]
\affiliation{
  \institution{University of Cambridge}
  \city{Cambridge}
  \country{UK}
}
\email{jh2298@cam.ac.uk}


\begin{abstract}
Deep learning has shown great promise in physiological signal analysis, yet its progress is hindered by heterogeneous data formats, inconsistent preprocessing strategies, fragmented model pipelines, and non-reproducible experimental setups. To address these limitations, we present Tyee, a unified, modular, and fully-integrated configurable toolkit designed for intelligent physiological healthcare. Tyee introduces three key innovations: (1) a \textbf{unified} data interface and \textbf{configurable} preprocessing pipeline for 12 kinds of signal modalities; (2) a \textbf{modular} and extensible architecture enabling flexible integration and rapid prototyping across tasks; and (3) \textbf{end-to-end}
workflow configuration, promoting reproducible and scalable experimentation. 
Tyee demonstrates consistent practical effectiveness and generalizability, outperforming or matching baselines across all evaluated tasks (with state-of-the-art results on 12 of 13 datasets).
The Tyee toolkit is released at ~\url{https://github.com/SmileHnu/Tyee} and actively maintained.
\end{abstract}




\maketitle

\section{Introduction}
\label{sec:intro}

\begin{table*}
\caption{{\textbf{Comparison of Tyee and several existing toolkits.} ``Code'' = only usable via code, ``Code/Config'' = both code and configuration files can be used, ``DL''= deep learning, \Checkmark = supported, \XSolidBrush = unsupported.}}
\label{tab:toolkit}
\begin{tabular}{lccccc}
\toprule
\textbf{Toolkit} & \textbf{Modality} & \textbf{Preprocessing} & \textbf{DL Support} & \textbf{Modular} & 
\textbf{Pipeline}\\
\midrule
MNE-Python~\cite{gramfort2013mne}      & EEG, MEG, ECoG, sEEG, fNIRS, EMG, EOG, ECG     & \makecell{Code} & \XSolidBrush & \XSolidBrush & \XSolidBrush \\
NeuroKit2~\cite{makowski2021neurokit2} & ECG, PPG, EDA, EMG, EOG, Resp, HRV & \makecell{Code} & \XSolidBrush & \Checkmark   & \XSolidBrush \\
PyHealth~\cite{pyhealth2023yang}       & EHR (ts, codes, tabular)                  & \makecell{Code} & \Checkmark   & \Checkmark   & Code   \\
TorchEEG~\cite{zhang2024torcheegemo}   & EEG only                                   & \makecell{Code} & \Checkmark   & \Checkmark   & Code \\
TorchECG~\cite{wen2022novel}           & ECG only                                    & \makecell{Code} & \Checkmark   & \Checkmark   & Code \\
\textbf{Tyee (Ours)} & \textbf{over 12 physiological signals} & \textbf{\makecell{Code/Config}} & \Checkmark & \Checkmark & \textbf{Code/Config} \\
\bottomrule
\end{tabular}
\end{table*}

Physiological signals, such as electroencephalography (EEG), electrocardiography (ECG), and electromyography (EMG), are essential to understand physiological and cognitive states~\cite{williams2023wearable, nie2015beyond}. Their growing utilization in intelligent healthcare has driven deep learning methods for physiological signal analysis~\cite{chen2022toward, han2021deep}.
In particular, these methods have shown promising performance in critical healthcare tasks from disease prediction to patient monitoring~\cite{ma2024sleepmg, yoo2023wireless}.

Recently, several open-source toolkits have been developed to support physiological signal analysis. As summarized in Table~\ref{tab:toolkit}, these toolkits provide a strong foundation and have contributed significantly to advancing the field.  However, despite their utility, researchers and developers (e.g., physiological signal engineers and machine learning practitioners) continue to face \textit{three key challenges} that limit broader adoption.
First, \textit{data integration remains a major challenge} due to fragmented data formats and inconsistent preprocessing pipelines across datasets; \textit{a simple, flexible interface} that hides these differences and makes it easier to work with data from many sources in a consistent way is still lacking.
Second, \textit{model development is often slowed} by rigid, monolithic toolkits that limit experimentation and make fair comparisons difficult; \textit{a modular framework} that encapsulates core components while enabling independent configuration and substitution remains an unmet need.
Third, \textit{inconsistent workflows, hidden defaults, and inflexible codebases} often lead to inconsistent and unreliable outcomes; \textit{a streamlined system that automates the pipeline}—reducing manual coding while maintaining transparency—remains critically needed.

To address these three challenges, we introduce an open-source \textit{\textbf{Tyee}: a unified, modular, and fully-integrated configurable deep learning \textbf{T}oolkit for ph\textbf{y}siological h\textbf{e}alth car\textbf{e}}. 
The name Tyee, pronounced like the Chinese term ``\begin{CJK}{UTF8}{gbsn}太医\end{CJK}'' (meaning \textit{imperial physician}), reflects its mission to bridge intelligent computing with physiological healthcare.
As the first toolkit of its kind, Tyee pioneers accessible physiological signal analysis. It is designed to serve both \textit{novice users} requiring ready-to-use pipelines and \textit{advanced researchers and developers} focused on developing, evaluating, or benchmarking custom models across diverse datasets and tasks. To accomplish this, it provides: (i) a unified data interface that abstracts dataset-specific preprocessing through configurable pipelines, enabling seamless integration and reuse; (ii) a modular architecture supporting flexible configuration (models, loss functions, optimizers, and evaluation metrics), enabling rapid prototyping and fair benchmarking; and (iii) holistic workflow configuration that automates the entire experimental process, from data preprocessing to evaluation, boosting reproducibility and scalability with minimal coding.

\section{Tyee: Design \& Features}
\label{sec:design}
In this section, we first present the overall architecture of Tyee, followed by a detailed discussion of its key features, including the unified data interface, configurable preprocessing, modular component design, and end-to-end configuration.

\begin{figure*}[!t]
  \centering
  \includegraphics[width=0.89\linewidth]{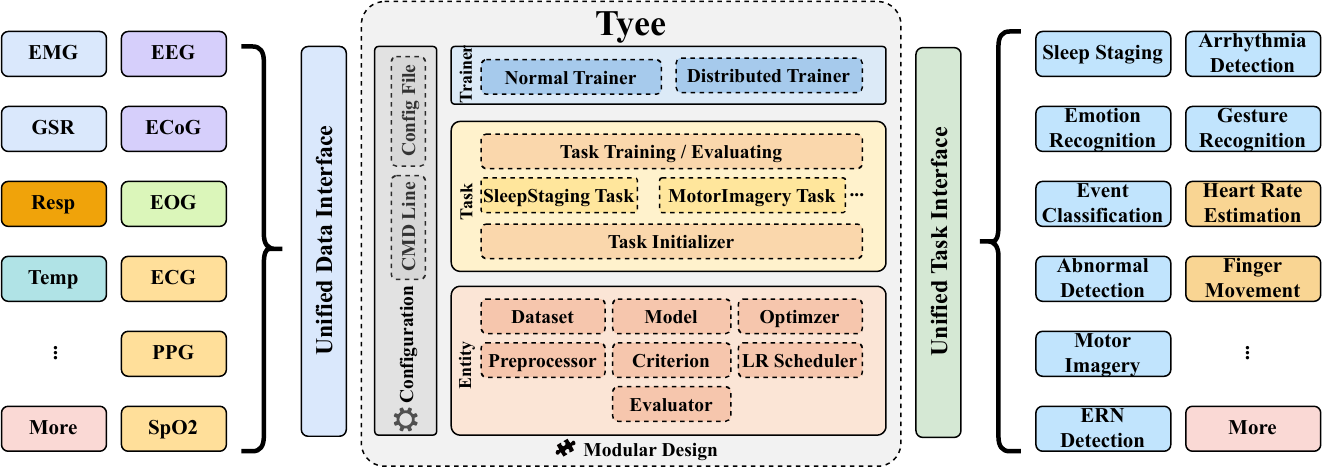}
  \caption{\textbf{Architecture of the Tyee toolkit.} It features modular components and end-to-end configurability, supporting a diverse range of physiological signals and healthcare applications via unified data and task interfaces.}
  \Description{Tyee toolkit architecture.}
  \label{fig:architecture}
\end{figure*}

\subsection{Design Principles \& Tyee Architecture} Inspired by prior works~\cite{yang2021superb, jarrettclairvoyance} and to address key limitations in existing physiological signal toolkits, Tyee is developed based on three guiding principles. \textit{\textbf{Reproducibility}} is improved by decomposing the workflow into modular components with encapsulated configurations, enabling transparent editing and reliable evaluation.
\textit{\textbf{Standardization}} is met by including frequently used modules with a single unified interface, lowering the barrier for collaborative and interdisciplinary research.
\textit{\textbf{Extensibility}} is achieved by encapsulating and integrating new models or algorithms via lightweight coding efforts, supporting flexible and scalable development.

Following these design principles, we present the architecture of Tyee in Figure~1. The system is composed of configurable modules structured within a layered architecture, enabling a flexible yet end-to-end workflow for physiological signal analysis. From a bottom-up perspective, it encapsulates all major processing steps across four layers—\textit{Entity, Task, Training, and Configuration}—with each layer abstracting a specific functionality.

\subsection{Unified Data Interface \& Preprocessing}
This design enables users to standardize diverse datasets and customize preprocessing workflows tailored to specific task requirements.
By encapsulating common data operations, such as reading, preprocessing, and storing, into a consistent and extensible workflow, Tyee simplifies handling heterogeneous physiological signals. Specifically, it provides the following core functionalities.
\begin{itemize}
    \item \texttt{read\_record()}: this allows users to load physiological signals from various file formats (e.\,g., .edf, .mat, .cnt, .bdf), and supports easy integration of new datasets through user-defined format-specific parsers.
    \item \texttt{online\_transform} and \texttt{offline\_transform}: two kinds of configurable transform parameters are introduced to accommodate varying preprocessing requirements.  This enables task-specific processing of signals and labels via configuration files, respectively. To further streamline development, it also offers a collection of built-in transforms covering common preprocessing routines.
    \item \texttt{\_\_getitem\_\_()}: this provides a unified interface for retrieving preprocessed samples, ensuring consistent and efficient access for various tasks.
\end{itemize}

\subsection{Modular Components}
Tyee adopts a layered architecture, including the Entity Layer, Task Layer, Training Layer, and Configuration Layer, as illustrated in Figure~\ref{fig:architecture}. Each layer is structured with well-defined interfaces and encapsulates modular components.

\subsubsection*{Entity Layer}
\label{sec:entity}
This encapsulates core modeling components, including:
datasets, models, criteria, optimizers, and evaluation metrics. 

\textit{Data Module} supports diverse physiological signals via a unified interface and configurable preprocessing. New datasets can be integrated by implementing the \texttt{read\_record()} method. Each data transform (called Preprocessor) is implemented as a subclass of the \texttt{BaseTransform} class and overrides the \texttt{transform()} method. Custom preprocessors can be registered and configured through YAML files.

\textit{Model Module} is built on PyTorch and incorporates a range of deep learning models commonly applied to physiological signal analysis, such as sleep staging, seizure detection, and emotion recognition. New models can be added by implementing standard PyTorch classes (i.\,e., subclasses of \texttt{torch.nn.Module}) using conventional \texttt{\_\_init\_\_()} and \texttt{forward()} methods. 

\textit{Optimizer and Other Modules.} Additional components such as optimizers, criteria, learning rate schedulers, and evaluators are encapsulated within the Entity Layer. Tyee includes a collection of built-in criteria, learning rate schedulers, and common evaluation metrics to support various training strategies and tasks.

\subsubsection*{Task Layer}
This layer dynamically loads and initializes components such as the model, dataset, criteria, and metrics based on module paths and class names specified in the configuration file. It orchestrates these components to execute training and evaluation workflows. Users can customize behavior by overriding \texttt{train\_step()} and \texttt{valid\_step()} methods. 

\subsubsection*{Training Layer}
This layer manages the common and end-to-end procedures for training and evaluation. It coordinates task initialization, task execution, checkpointing, and logging. To meet the demands of large-scale computational resources, the Training Layer supports distributed training via standard PyTorch strategies.

\subsubsection*{Configuration Layer}
This layer provides centralized environment settings through YAML configuration files. It governs the setup of datasets, models, tasks, and hyper-parameters. Additionally, Tyee also provides a unified command-line interface to enable fast overrides of configuration parameters.

\subsection{Configurable Pipeline}
Tyee adopts a YAML file-based configuration system that declaratively defines all experiment parameters through structured files. 
It supports consistent module initialization and management across experiments.
Each specific task (as shown in Figure~1 right) is defined within a YAML file, outlining all components of the experiment pipeline in logic sections, from data handling to model architecture and training routines.
More specifically, it consists of \textit{common} for global settings, \textit{dataset} for data access, \textit{model} for model details, \textit{optimizer} for learning strategies, \textit{task} for training objective, \textit{trainer} for training-specific configurations, and \textit{distributed} for distributed training setup.

During training, the YAML file is loaded, parsed and passed to the task layer to initialize corresponding components.
Users can fully define, reproduce, and adapt experiments by simply modifying configuration files, without changing the underlying code. 
In addition, Tyee also provides a command-line interface to launch workflows using specified YAML files or runtime overrides.

Complete documentation of all configurable parameters is available in the GitHub repository.

\section{Evaluation and Results}

\subsubsection*{Datasets \& Models \& Metrics}
\label{sec:datasets}

\begin{table*}[!ht]
\caption{\textbf{Overview of the 13 datasets used to benchmark Tyee, covering 12 physiological signal types, 11 baseline models, and 11 different healthcare tasks.} Top: eight unimodal datasets. Bottom: five multimodal ones. Statistics include subject count, sampling rate (SR), sample size, and evaluation metrics. \textit{Full dataset references and original model implementations are available on GitHub.} }

\label{tab:datasets}
\small
\begin{tabular}{ccccrrrc}
\toprule
\textbf{Dataset} & \textbf{Task} & \textbf{Signals} & \textbf{Model} & \textbf{\#Subj.} & \textbf{SR (Hz)} & \textbf{\#Samples} & \textbf{Evaluation Metrics}  \\
\midrule
TUEV       & Event Classification & EEG                       & LaBraM          & 2,383 & 256   & 112,491    & BA, WF1, $\kappa$ \\
TUAB       & Abnormal Detection   & EEG                       & LaBraM          & 288   & 256   & 409,455    & BA, AUPRC, AUROC  \\
BCIC-2A    & Motor Imagery        & EEG                       & EEGConformer    & 9     & 250   & 5,760      & Accuracy          \\
BCIC-4     & Finger Movement      & ECoG                      & FingerFlex      & 3     & 1,000 & 39,726     & Mean CC           \\
KaggleERN  & ERN Detection        & EEG                       & EEGPT           & 26    & 200   & 8,840      & BA, AUROC, $\kappa$\\
PhysioP300 & P300 Recognition     & EEG                       & EEGPT           & 12    & 2,048 & 23,699     & BA, AUROC, $\kappa$\\
MIT-BIH    & Arrhythmia Detection & ECG                       & ECGResNet34     & 40    & 360   & 103,979    & Accuracy           \\
NinaproDB5 & Gesture Recognition   & EMG                       & EMGBench        & 10    & 200   & 39,597     & Accuracy            \\
\midrule
PPG-DaLiA  & HR Estimation        & PPG, 3D Acc               & BeliefPPG       & 15    & 64/32 & 64,607     & MAE                \\
SEED-V     & Emotion Recognition  & EEG, EOG                  & G2G-ResNet18    & 20    & 1,000 & 148,080   & Accuracy            \\
DEAP       & Emotion Recognition  & GSR, PPG, Resp, Temp      & MLSTM-FCN       & 32    & 512   & 25,600    & Accuracy            \\
SleepEDFx  & Sleep Staging        & EEG, EOG                  & SalientSleepNet & 20    & 100   & 42,037    & Accuracy, F1         \\
CinC2018   & Sleep Staging        & EEG, EOG, ECG, SaO2, Resp & SleepFM         & 1,985 & 200   & 221,939   & F1, AUROC, AUPRC    \\
\bottomrule
\end{tabular}
\end{table*}

To demonstrate the practical utility and versatility of Tyee, we conducted extensive experiments using a diverse collection of physiological signal datasets. We included comprehensive examples for our evaluation, as detailed in Table~\ref{tab:datasets}. In particular, we systematically evaluated the toolkit on 13 public datasets, covering 12 distinct signal types and 11 tasks. The evaluation features state-of-the-art models, including the recent EEGPT. In total, over 4,800 subjects and their 1.2M+ samples were involved, and the supported signal sampling rates range from 32 Hz to over 2 kHz, demonstrating Tyee's flexibility, with actual capability likely extending beyond these benchmarks.

For \textit{unimodal} tasks, we evaluated eight benchmark datasets. Advanced models from prior studies were reproduced using original implementations or publicly available code, with necessary adaptations for compatibility within Tyee. For \textit{multimodal} tasks, we utilized five public datasets to assess Tyee's ability to process and integrate multiple physiological signals. Notably, the SleepEDFx dataset includes two subsets: SleepEDFx-39 and SleepEDFx-158, and here SleepEDFx-39 was used in our experiments. Again, relevant models were re-implemented to ensure compatibility within Tyee.
The only exception is the SleepFM model, which was deployed for the sleep staging task on CinC2018 dataset. The original logistic regression classifier was replaced with a linear head to ensure compatibility with the Tyee interface. This modification does not alter the model's core architecture or performance characteristics but allows for consistent handling of model outputs across tasks.

For the evaluation, in all cases, we followed the original studies' hyperparameter configurations and reported metrics as applied in their official implementations. This ensured fair and consistent comparisons at the most. In addition, to maintain uniformity in result presentation, we averaged metric values when multiple evaluation metrics were reported for a dataset. 
As an example, in the case of the SleepEDFx dataset, the original study reported both accuracy and F1-score; hence we reported the average of the two for consistency. Moreover, for the PPG-DaLiA dataset, we further transformed the MAE values using the formula $100 / \ln(1 + \text{MAE})$ to facilitate consistent scaling and comparability with other tasks.
The full catalog of the explored dataset references and original model implementations can be found in the project's GitHub repository, due to space limitations.

\begin{figure}[!ht]
  \centering
  \includegraphics[width=\linewidth, trim=0 12 0 0, clip]{radar.pdf}
  \caption{\textbf{Radar plots comparing Tyee (red) vs. baselines (blue) in (a) unimodal and (b) multimodal healthcare tasks.} Tyee matches or outperforms the baselines on 12 out of 13 datasets, demonstrating reliable and consistent benchmark reproduction. }
  \Description{The Tyee's performance of unimodal (left) and multimodal (right) on various tasks.}
  \label{fig:radar}
\end{figure}

\subsubsection*{Results on Unimodal Tasks}
\label{sec:uni-modal-eval}

Figure~\ref{fig:radar} presents performance comparisons between Tyee and official baseline models using radar plots for both (a) unimodal and (b) multimodal evaluations. As shown in Figure~\ref{fig:radar} (a), Tyee demonstrates comparable or superior performance across all eight unimodal tasks, with particularly strong results on the PhysioP300 and BCIC-4 datasets. While conservative model choices may underrepresent its full potential, these results highlight Tyee's ability to reliably reproduce benchmarks while maintaining robustness and generalizability across diverse physiological signal analysis tasks.

\subsubsection*{Results on Multimodal Tasks}
\label{sec:multi-modal-eval}
We observe a similar pattern of performance in the multimodal tasks, as shown in Figure~\ref{fig:radar} (b). Specifically, Tyee demonstrates performance on par with the official implementations, achieving slightly better results on three datasets (SEED-V, SleepEDFx, and DEAP), a substantial improvement on DEAP, and a slightly weaker (but not significantly) performance on PPG-DaLiA. Overall, these results highlight Tyee's efficiency and effectiveness in supporting multimodal physiological signal analysis.
\section{Conclusion}
\label{sec:conclusion}
We present Tyee, a user-friendly and comprehensive open-source toolkit for intelligent physiological signal analysis. It is released under the CC BY-NC 4.0 License and available on GitHub. We provide complete source code, detailed documentation, and illustrative examples spanning a wide range of physiological signals and healthcare applications, with easy installation via Conda and Docker. We envision Tyee as a foundational platform that accelerates research, fosters collaboration, and streamlines the innovation pipeline from hypothesis to deployment. By reducing redundancy and enhancing reproducibility, Tyee aims to facilitate the development of scalable intelligent health solutions with potential benefits for patient care and public health.

\begin{acks}
This work was supported by the Guangdong Basic and Applied Basic Research Foundation under Grant No.~2024A1515010112, and the Changsha Science and Technology Bureau Foundation under Grant No.~kq2402082. 
\end{acks}

\bibliographystyle{ACM-Reference-Format}
\balance
\bibliography{bibliography}





\end{document}